\pdfoutput=1
\DeclareUnicodeCharacter{FB01}{fi}
\documentclass[11pt]{article}

\usepackage[]{acl}

\usepackage{times}
\usepackage{latexsym}

\usepackage[T1]{fontenc}

\usepackage[utf8]{inputenc}

\usepackage{microtype}
\usepackage{amsmath}
\usepackage{graphicx}
\usepackage{color}
\usepackage{multirow}
\usepackage{float}
\usepackage{diagbox}
\usepackage{array}
\usepackage{booktabs}
\usepackage{hhline}
\usepackage{makecell}
\usepackage{caption, subcaption}
\usepackage{tabularx}
\usepackage{booktabs}
\usepackage{color}
\usepackage[linesnumbered,ruled,vlined]{algorithm2e}

\title{Can Prompt Probe Pretrained Language Models? Understanding the Invisible Risks from a Causal View}

\author{Boxi Cao${}^{1,3}$, Hongyu Lin${}^{1}$, Xianpei Han${}^{1, 2,4}$\thanks{~ Corresponding Authors}, Fangchao Liu${}^{1,3}$, Le Sun${}^{1,2}$\footnotemark[1]\\
${}^{1}$Chinese Information Processing Laboratory ~ ${}^{2}$State Key Laboratory of Computer Science \\
Institute of Software, Chinese Academy of Sciences, Beijing, China\\
${}^{3}$University of Chinese Academy of Sciences, Beijing, China \\
${}^{4}$ Beijing Academy of Artificial Intelligence, Beijing, China\\
{\tt \{boxi2020,hongyu,xianpei,fangchao2017,sunle\}@iscas.ac.cn}}

\begin{document}
\maketitle
\newcommand{\M}{causalEval}
\begin{abstract}
Prompt-based probing has been widely used in evaluating the abilities of pretrained language models (PLMs). Unfortunately, recent studies have discovered such an evaluation may be inaccurate, inconsistent and unreliable. Furthermore, the lack of understanding its inner workings, combined with its wide applicability, has the potential to lead to unforeseen risks for evaluating and applying PLMs in real-world applications. To discover, understand and quantify the risks, this paper investigates the prompt-based probing from a causal view, highlights three critical biases which could induce biased results and conclusions, and proposes to conduct debiasing via causal intervention. This paper provides valuable insights for the design of unbiased datasets, better probing frameworks and more reliable evaluations of pretrained language models. Furthermore, our conclusions also echo that we need to rethink the criteria for identifying better pretrained language models\footnote{We openly released the source code and data at \url{https://github.com/c-box/causalEval}.}.

\end{abstract}

\section{Introduction}
\label{sec:intro}

During the past few years, the great success of pretrained language models (PLMs)~\citep{devlinBERTPretrainingDeep2019,liuRoBERTaRobustlyOptimized2019,brownLanguageModelsAre2020,DBLP:journals/jmlr/RaffelSRLNMZLL20} raises extensive attention about evaluating what knowledge do PLMs actually entail. 
One of the most popular approaches is prompt-based probing~\citep{petroniLanguageModelsKnowledge2019,feldmanCommonsenseKnowledgeMining2019,brownLanguageModelsAre2020,schickFewShotTextGeneration2020,ettingerWhatBERTNot2020,DBLP:journals/corr/abs-2107-02137}, 
which assesses whether PLMs are knowledgable for a specific task by querying PLMs with task-specific prompts.
For example, to evaluate whether BERT knows the birthplace of Michael Jordan, we could query BERT with ``Michael Jordan was born in [MASK]''. 
Recent studies often construct prompt-based probing datasets, and take PLMs’ performance on these datasets as their abilities for the corresponding tasks. 
Such a probing evaluation has been wildly used in many benchmarks such as SuperGLUE~\citep{DBLP:conf/nips/WangPNSMHLB19,brownLanguageModelsAre2020}, LAMA~\citep{petroniLanguageModelsKnowledge2019}, oLMpics~\citep{Talmor2020oLMpicsOnWL}, LM diagnostics~\citep{ettingerWhatBERTNot2020}, CAT~\citep{Zhou2020EvaluatingCI}, X-FACTR~\citep{jiangXFACTRMultilingualFactual2020}, BioLAMA~\citep{sung2021can}, etc.

\begin{figure}[!tp]
 \setlength{\belowcaptionskip}{-0.4cm}
  \centering
  \includegraphics[width=0.5\textwidth]{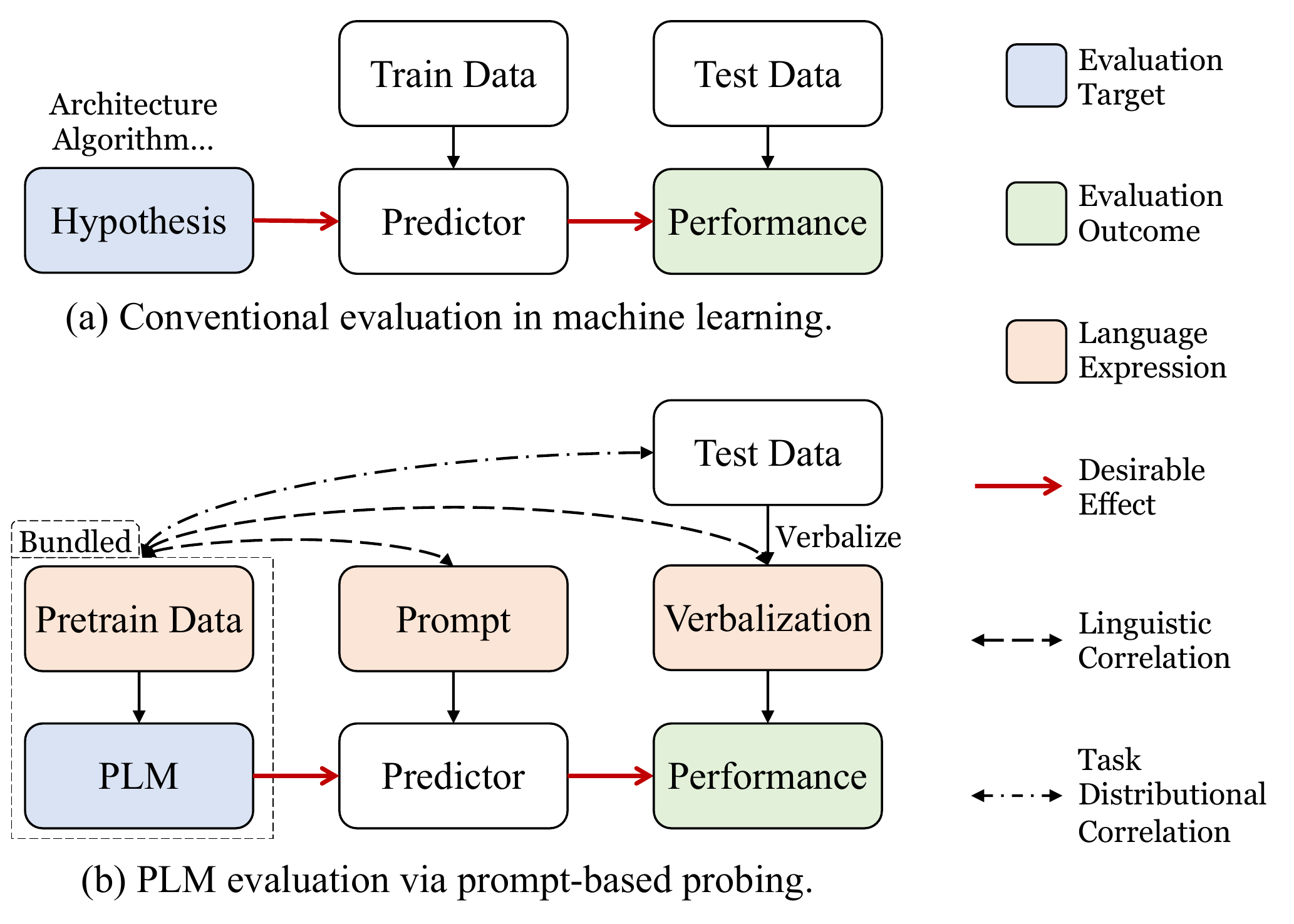}
  \caption{The illustrated procedure for two kinds of evaluation criteria.}
  \label{fig:head}

\end{figure}

Unfortunately, recent studies have found that evaluating PLMs via prompt-based probing could be inaccurate, inconsistent, and unreliable. For example,
~\citet{poernerEBERTEfficientYetEffectiveEntity2020} finds that the performance may be overestimated because many instances can be easily predicted by only relying on surface form shortcuts. 
~\citet{elazar2021measuring} shows that semantically equivalent prompts may result in quite different predictions.
~\citet{cao-etal-2021-knowledgeable} demonstrates that PLMs often generate unreliable predictions which are prompt-related but not knowledge-related.

In these cases, the risks of blindly using prompt-based probing to evaluate PLMs, without understanding its inherent vulnerabilities, are significant. Such biased evaluations will make us overestimate or underestimate the real capabilities of PLMs, mislead our understanding of models, and result in wrong conclusions. Therefore, to reach a trustworthy evaluation of PLMs, it is necessary to dive into the probing criteria and understand the following two critical questions: \emph{1) What biases exist in current evaluation criteria via prompt-based probing? 2) Where do these biases come from?}

To this end, we compared PLM evaluation via prompt-based probing with conventional evaluation criteria in machine learning. 
Figure~\ref{fig:head} shows their divergences.
Conventional evaluations aim to evaluate different hypotheses (e.g., algorithms or model structures) for a specific task. The tested hypotheses are raised independently of the training/test data generation.
However, this independence no longer sustains in prompt-based probing.
There exist more complicated implicit connections between pretrained models, probing data, and prompts, mainly due to the bundled pretraining data with specific PLMs.
These unaware connections serve as invisible hands that can even dominate the evaluation criteria from both linguistic and task aspects.
From the linguistic aspect, because pretraining data, probing data and prompts are all expressed in the form of natural language, there exist inevitable \emph{linguistic correlations} which can mislead evaluations.
From the task aspect, the pretraining data and the probing data are often sampled from correlated distributions. Such invisible \emph{task distributional correlations} may significantly bias the evaluation. For example, Wikipedia is a widely used pretraining corpus, and many probing data are also sampled from Wikipedia or its extensions such as Yago, DBPedia or Wikidata~\citep{petroniLanguageModelsKnowledge2019, jiangXFACTRMultilingualFactual2020,sung2021can}. 
As a result, such task distributional correlations will inevitably confound evaluations via domain overlapping, answer leakage, knowledge coverage, etc.

\begin{figure*}[htp]
  \setlength{\belowcaptionskip}{-0.4cm}
  \centering
  \includegraphics[width=\textwidth]{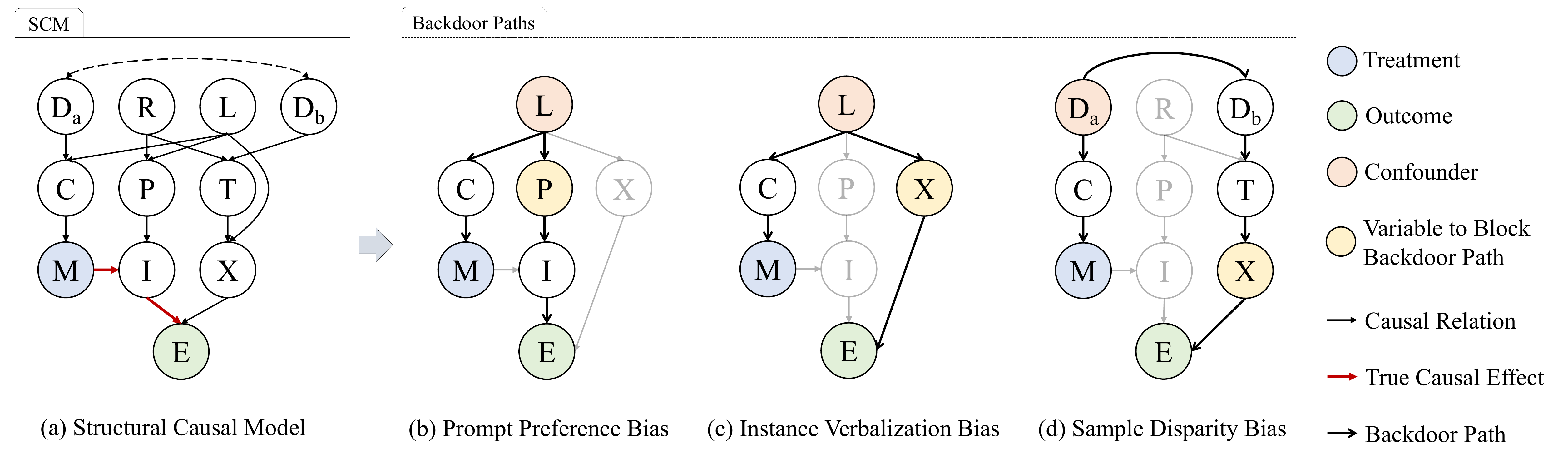}
  \caption{The structural causal model for factual knowledge probing and the three backdoor paths in SCM correspond to three biases.}
  \label{fig:detail_scm}
\end{figure*}

To theoretically identify how these correlations lead to biases, we revisit the prompt-based probing from a causal view.
Specifically, we describe the evaluation procedure using a structural causal model~\citep{pearl2000models} (SCM),  which is shown in Figure~\ref{fig:detail_scm}a.
Based on the SCM, we find that the linguistic correlation and the task distributional correlation correspond to three backdoor paths in Figure~\ref{fig:detail_scm}b-d, which lead to three critical biases:
\begin{itemize}
  \item \textbf{Prompt Preference Bias,} which mainly stems from the underlying linguistic correlations between PLMs and prompts, i.e., the performance may be biased by the fitness of a prompt to PLMs’ linguistic preference. For instance, semantically equivalent prompts will lead to different biased evaluation results.

  \item \textbf{Instance Verbalization Bias,} which mainly stems from the underlying linguistic correlations between PLMs and verbalized probing datasets, i.e., the evaluation results are sensitive and inconsistent to the different verbalizations of the same instance (e.g., representing the U.S.A. with the U.S. or America).

  \item \textbf{Sample Disparity Bias,} which mainly stems from the invisible distributional correlation between pretraining and probing data, i.e., the performance difference between different PLMs may due to the sample disparity of their pretraining corpus, rather than their ability divergence. Such invisible correlations may mislead evaluation results, and thus lead to implicit, unaware risks of applying PLMs in real-world applications.
\end{itemize}

We further propose to conduct causal intervention via backdoor adjustments, which can reduce bias and ensure a more accurate, consistent and reliable probing under given assumptions. Note that this paper not intends to create a ``universal correct'' probing criteria, but to remind the underlying invisible risks, to understand how spurious correlations lead to biases, and to provide a causal toolkit for debiasing probing under specific assumptions. 
Besides, we believe that our discoveries not only exist in prompt-based probing, but will also influence all prompt-based applications to pretrained language models. Consequently, our conclusions echo that we need to rethink the criteria for identifying better pretrained language models with the above-mentioned biases.

Generally, the main contributions of this paper are:
\begin{itemize}
  \item We investigate the critical biases and quantify their risks of evaluating pretrained language models with widely used  prompt-based probing, including prompt preference bias, instance verbalization bias, and sample disparity bias.
  \item We propose a causal analysis framework, which can be used to effectively identify, understand, and eliminate biases in prompt-based probing evaluations.
  \item We provide valuable insights for the design of unbiased datasets, better probing frameworks, and more reliable evaluations, and echo that we should rethink the evaluation criteria for pretrained language models.
\end{itemize}

\section{Background and Experimental Setup}
\subsection{Causal Inference}
\label{ssec:causal}
Causal inference is a promising technique for identifying undesirable biases and fairness concerns in benchmarks~\citep{DBLP:conf/nips/HardtPNS16, DBLP:conf/nips/KilbertusRPHJS17,DBLP:conf/nips/KusnerLRS17,Vig2020InvestigatingGB,feder2021causal}. Causal inference usually describes the causal relations between variables via Structural Causal Model (SCM), then recognizes confounders and spurious correlations for bias analysis, finally identifies true causal effects by eliminating biases using causal intervention techniques.

\paragraph{SCM} The structural causal model~\citep{pearl2000models} describes the relevant features in a system and how they interact with each other. Every SCM is associated with a graphical causal model $G = \{V, f\}$, which consists of a set of nodes representing variables $V$, as well as a set of edges between the nodes representing the functions $f$ to describe the causal relations.

\paragraph{Causal Intervention} To identify the true causal effects between an ordered pair of variables $(X, Y)$, Causal intervention fixes the value of $X=x$ and removes the correlations between $X$ and its precedent variables, which is denoted as $do(X=x)$. In this way,  $\mathcal{P}(Y=y|do(X=x))$ represents the true causal effects of treatment $X$ on outcome $Y$~\citep{pearl2016causal}.

\paragraph{Backdoor Path} When estimating the causal effect of $X$ on $Y$, the backdoor paths are the non-causal paths between $X$ and $Y$ with an arrow into $X$, e.g., $X \leftarrow Z \rightarrow Y$. Such paths will confound the effect that $X$ has on $Y$ but not transmit causal influences from $X$, and therefore introduce spurious correlations between $X$ and $Y$.

\paragraph{Backdoor Criterion} The Backdoor Criterion is an important tool for causal intervention. 
Given an ordered pair of variables $(X, Y)$ in SCM, and a set of variables $Z$ where $Z$ contains no descendant of $X$ and blocks every backdoor path between $X$ and $Y$, then the causal effects of $X=x$ on $Y$ can be calculated by:
\begin{equation}
    \begin{split}
    & \mathcal{P}(Y=y|do(X=x)) = \\ & \sum_{z}\mathcal{P}(Y=y|X=x, Z=z)\mathcal{P}(Z=z),
    \end{split}
\end{equation}
where $\mathcal{P}(Z=z)$ can be estimated from data or priorly given, and is independent of $X$.

\subsection{Experimental Setup}
\paragraph{Task} 
This paper investigates prompt-based probing on one of the most representative and well-studied tasks -- factual knowledge probing~\citep{liu2021pre}. For example, to evaluate whether BERT knows the birthplace of Michael Jordan, factual knowledge probing queries BERT with ``Michael Jordan was born in [MASK]'', where \texttt{Michael Jordan} is the verbalized subject mention, \textit{``was born in''} is the verbalized prompt of relation \texttt{birthplace}, and [MASK] is a placeholder for the target object.

\paragraph{Data} We use LAMA~\citep{petroniLanguageModelsKnowledge2019}
as our primary dataset, which is a set of knowledge triples sampled from Wikidata. We remove the N-M relations~\citep{elazar2021measuring} which are unsuitable for the P@1 metric and retain 32 probing relations in the dataset. Please refer to the appendix for detail.

\paragraph{Pretrained Models} We conduct probing experiments on 4 well-known PLMs: BERT~\citep{devlinBERTPretrainingDeep2019}, RoBERTa~\citep{liuRoBERTaRobustlyOptimized2019}, GPT-2~\citep{radfordLanguageModelsAre2019} and BART~\citep{lewis-etal-2020-bart}, which correspond to 3 representative PLM architectures, including autoencoder (BERT, RoBERTa), autoregressive (GPT-2) and denoising autoencoder (BART).

\section{Structural Causal Model for Factual Knowledge Probing}
\label{sec:scm}

In this section, we formulate the SCM for factual knowledge probing procedure and describe the key variables and causal relations.

The SCM is shown in Figure~\ref{fig:detail_scm}a, which contains 11 key variables:
1) \textbf{Pretraining corpus distribution} $D_a$;
2) \textbf{Pretraining corpus} $C$, e.g., Webtext for GPT2, Wikipedia for BERT;
3) \textbf{Pretrained language model} $M$;
4) \textbf{Linguistic distribution} $L$, which guides how a concept is verbalized into natural language expression, e.g., relation to prompt, entity to mention;
5) \textbf{Relation} $R$, e.g., \texttt{birthplace}, \texttt{capital}, each relation corresponds to a probing task;
6) \textbf{Verbalized prompt} $P$ for each relation , e.g, \textit{$x$ was born in $y$};
7) \textbf{Task-specific predictor} $I$, which is a PLM combined with a prompt, e.g., <BERT, \textit{was born in}> as a \texttt{birthplace} predictor;
8) \textbf{Probing data distribution} $D_b$, e.g., fact distribution in Wikidata;
9) \textbf{Sampled probing data} $T$ such as LAMA, which are sampled entity pairs (e.g., <Q41421, Q18419> in Wikidata) of relation $R$;
10) \textbf{Verbalized instances} $X$, (e.g., <Michael Jordan, Brooklyn> from <Q41421, Q18419>);
11) \textbf{Performance} $E$ of the predictor $I$ on $X$.

The causal paths of the prompt-based probing evaluation contains:
\begin{itemize}
  \item \textbf{PLM Pretraining.} The path $\{D_a, L\} \rightarrow C \rightarrow M$ represents the pretraining procedure for language model $M$,
  which first samples pretraining corpus $C$ according to pretraining corpus distribution $D_a$ and linguistic distribution $L$, then pretrains $M$ on $C$.
  \item \textbf{Prompt Selection.} The path $\{R, L\} \rightarrow P$ 
  represents the prompt selection procedure, where each prompt $P$ must exactly express the semantics of relation $R$, and will be influenced by the linguistic distribution $L$.
  \item \textbf{Verbalized Instances Generation.} The path $\{D_b, R\} \rightarrow T \rightarrow X \leftarrow L$  represents the generation procedure of verbalized probing instances $X$, which first samples probing data $T$ of relation $R$  according to data distribution $D_b$, then verbalizes the sampled data $T$ into $X$ according to the linguistic distribution $L$.
  \item \textbf{Performance Estimation.} The path $\{M, P\} \rightarrow I \rightarrow E \leftarrow X$ represents the performance estimation procedure, where the predictor $I$ is first derived by combining PLM $M$ and prompt $P$, and then the performance $E$ is estimated by applying predictor $I$ on verbalized instances $X$.
\end{itemize}

To evaluate PLMs’ ability on fact extraction, we need to estimate $\mathcal{P}(E|do(M=m), R=r)$. 
Such true causal effects are represented by the path $M \rightarrow I \rightarrow E$ in SCM. 
Unfortunately, 
there exist three backdoor paths between pretrained language model $M$ and performance $E$, as shown in Figure~\ref{fig:detail_scm}b-d.
These spurious correlations make the observation correlation between $M$ and $E$ cannot represent the true causal effects of $M$ on $E$, and will inevitably lead to biased evaluations.
In the following, we identify three critical biases in the prompt-based probing evaluation and describe the manifestations, causes, and casual interventions for each bias.

\section{Prompt Preference Bias}
In prompt-based probing, the predictor of a specific task (e.g., the knowledge extractor of relation \texttt{birthplace}) is a PLM $M$ combined with a prompt $P$ (e.g., BERT + \textit{was born in}). 
However, PLMs are pretrained on specific text corpus, therefore will inevitably prefer prompts sharing the same linguistic regularity with their pretraining corpus. 
Such implicit prompt preference will confound the true causal effects of PLMs on evaluation performance, i.e., the performance will be affected by both the task ability of PLMs and the preference fitness of a prompt. In the following, we investigate prompt preference bias via causal analysis.

\begin{figure}[tp]
  \centering
  \includegraphics[width=0.48\textwidth]{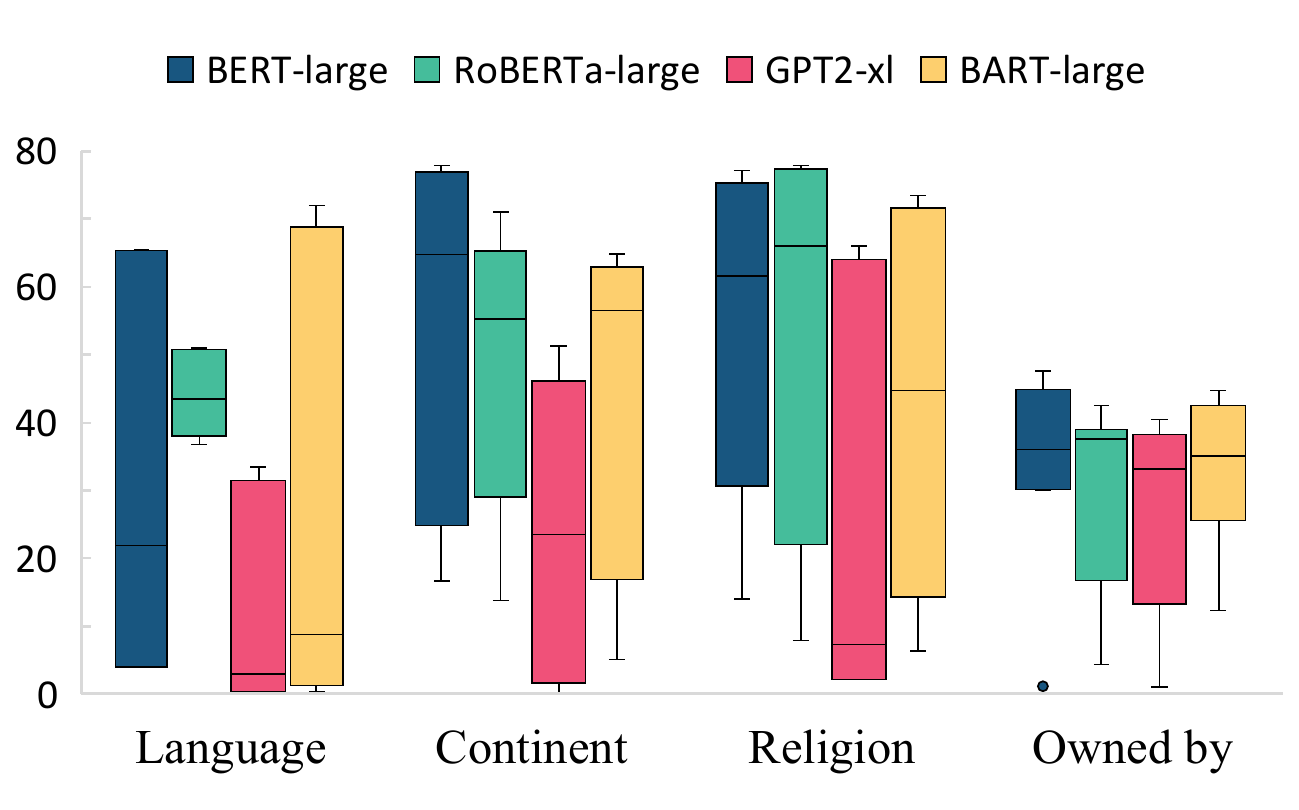}
  \caption{The variances of P@1 performance of 4 PLMs on 4 relations using semantically equivalent prompts. We can see the performance varies significantly. }
  \label{fig:prompt_box}
\end{figure}

\subsection{Prompt Preference Leads to Inconsistent Performance}
\label{ssec:utterance_affect}
In factual knowledge probing, we commonly assign one prompt for each relation (e.g., \texttt{X was born in Y} for \texttt{birthplace}). 
However, different PLMs may prefer different prompts, and it is unable to disentangle the influence of prompt preference from the final performance. Such invisible prompt preference will therefore lead to inconsistent conclusions.

To demonstrate this problem, we report the performance variance on LAMA using different prompts for each PLM. For each relation, we follow ~\citet{elazar2021measuring,jiangHowCanWe2020} and design at least 5 prompts that are semantically equivalent and faithful but vary in linguistic expressions.

\textbf{Prompt selection significantly affects performance.} Figure~\ref{fig:prompt_box} illustrates the performance on several relations, where the performances of all PLMs vary significantly on semantically equivalent prompts.
For instance, by using different prompts, the Precision@1 of relation \texttt{languages spoken} dramatically changing from 3.90\% to 65.44\% on BERT-large, and from 0.22\% to 71.94\% on BART-large.
This result is shocking, because the same PLM can be assessed from ``knowing nothing'' to ``sufficiently good'' by only changing its prompt. 
Table~\ref{tab:prompt_var} further shows the quantitative results, 
for BERT-large, the averaged standard deviation of Precision@1 of different prompts is 8.75.
And the prompt selection might result in larger performance variation than model selection:
on more than 70\% of relations, the best and worst prompts will lead to >10 point variation at Precision@1, which is larger than the majority of performance gaps between different models. 

\begin{figure}[tp]
  \centering
  \includegraphics[width=0.48\textwidth]{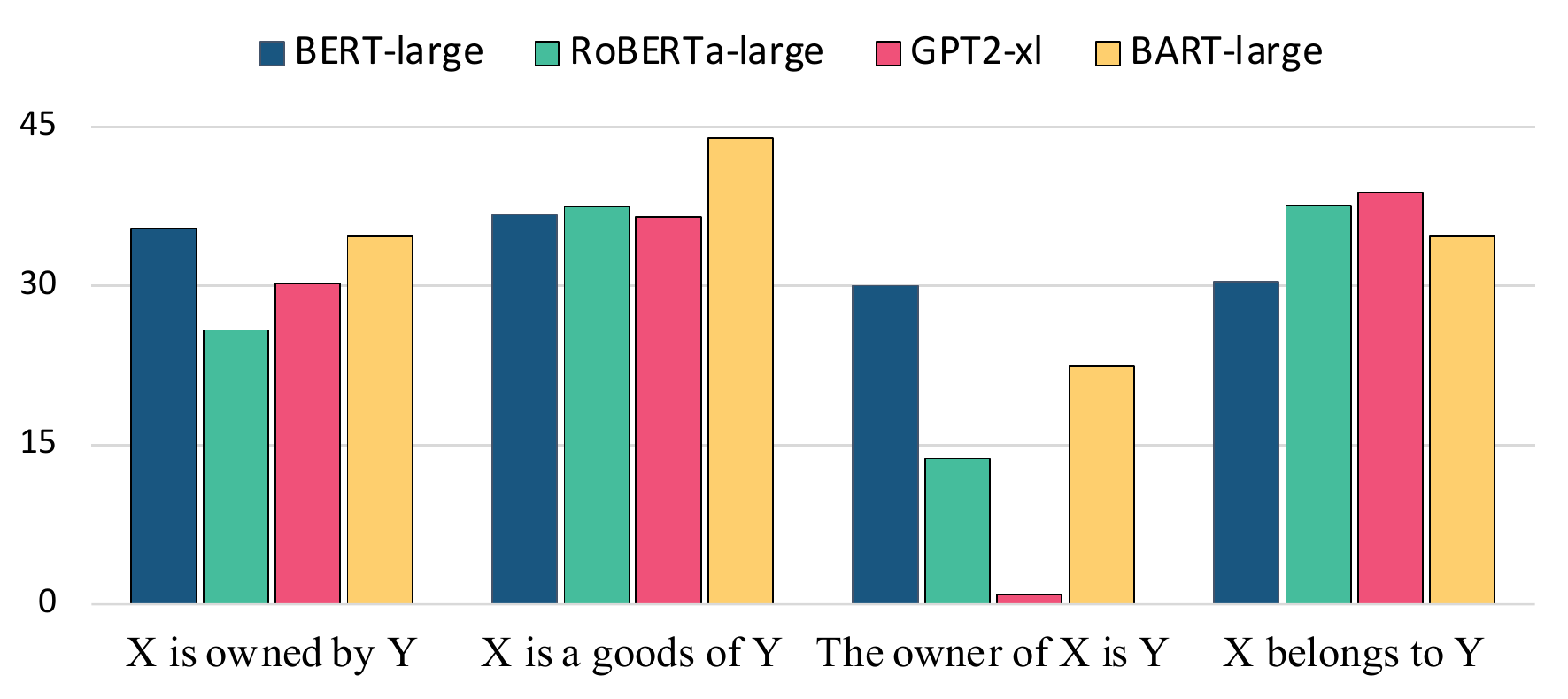}
  \caption{The P@1 performance of 4 PLMs using 4 different prompts of relation \texttt{owned by}, where the rank of 4 PLMs is unstable on different prompts: prompt preference leads to 3 distinct “best” models and 3 distinct “worst” models.
  }
  \label{fig:prompt_bar}
\end{figure}

\textbf{Prompt preference also leads to inconsistent comparisons.} Figure~\ref{fig:prompt_bar} demonstrates an example, where the ranks of PLMs are significantly changed when applying diverse prompts.
We also conduct quantitative experiments, which show that the PLMs' ranks on 96.88\% relations are unstable when prompt varies.

All these results demonstrate that the prompt preference bias will result in inconsistent performance. Such inconsistent performance will further lead to unstable comparisons between different PLMs, and therefore significantly undermines the evaluations via prompt-based probing.

\begin{table}[tp]
\setlength{\belowcaptionskip}{-0.4cm}
\resizebox{\columnwidth}{!}{
\begin{tabular}{lcccc}
\toprule
\textbf{Models}  & \textbf{LAMA P@1} & \textbf{Worst P@1} & \textbf{Best P@1} & \textbf{Std} \\ \hline
BERT-large       & 39.08                 & 23.45              & 46.73             & 8.75      \\     
RoBERTa-large    & 32.27                & 15.64              & 41.35             & 9.07       \\
GPT2-xl          & 24.19                 & 11.19              & 33.52             & 8.56       \\
BART-large       & 27.68                 & 16.21              & 38.93             & 8.35       \\ 
\bottomrule
\end{tabular}
}
\caption{The P@1 performance divergence of prompt selection averaged over all relations, we can see prompt preference results in inconsistent performance.}
\label{tab:prompt_var}
\end{table}

\subsection{Cause of Prompt Preference Bias}
\label{ssec:cause_utterance}

Figure~\ref{fig:detail_scm}b shows the cause of the prompt preference bias. 
When evaluating the ability of PLMs on specific tasks,
we would like to measure the causal effects of path $M \rightarrow I \rightarrow E$. However, because the prompt $P$ and the PLM $M$ are all correlated to the linguistic distribution $L$, there is a backdoor path $M \leftarrow C \leftarrow L \rightarrow P \rightarrow I \rightarrow E$ between PLM $M$ and performance $E$. Consequently, the backdoor path will confound the effects of $M \to I \to E$ with $P \to I \to E$.

Based on the above analysis, the prompt preference bias can be eliminated by blocking this backdoor path via backdoor adjustment, which requires a prior formulation of the distribution $\mathcal{P}(P)$. In Section~\ref{sec:reduce}, we will present one possible causal intervention formulation which can lead to more consistent evaluations.

\section{Instance Verbalization Bias}
Apart from the prompt preference bias, the underlying linguistic correlation can also induce bias in the instance verbalization process. Specifically, an instance in probing data can be verbalized into different natural language expressions (e.g., verbalize \texttt{Q30} in Wikidata into \texttt{America} or \texttt{the U.S.}), and different PLMs may prefer different verbalizations due to mention coverage, expression preference, etc. This will lead to instance verbalization bias.

\subsection{Instance Verbalization Brings Unstable Predictions}

In factual knowledge probing, each entity is verbalized to its default name. However, different PLMs may prefer different verbalizations, and such underlying correlation is invisible. Because we couldn’t measure how this correlation affects probing performance, the evaluation may be unstable using different verbalizations.

Table~\ref{tab:sub_var} shows some intuitive examples. When we query BERT ``The capital of the U.S. is [MASK]'', the answer is \texttt{Washington}. Meanwhile, BERT would predict \texttt{Chicago} if we replace  \texttt{the U.S.} to its alias \texttt{America}. 
Such unstable predictions make us unable to obtain reliable conclusions on whether or to what degree PLMs actually entail the knowledge.

\begin{table}[tp]
  \centering
  \resizebox{0.8\columnwidth}{!}{
  \begin{tabular}{lll}
  \toprule
  \textbf{Relation}           & \textbf{Mention}          & \textbf{Prediction} \\ \hline
  \multirow{4}{*}{Capital of} & America          & Chicago    \\
                              & the U.S.         & Washington \\ \cline{2-3} 
                              & China            & Beijing    \\
                              & Cathay           & Bangkok    \\ \hline
  \multirow{4}{*}{Birthplace} & Einstein         & Berlin     \\
                              & Albert Einstein  & Vienna     \\  \cline{2-3} 
                              
                              & Isaac Newton     & London     \\
                              & Sir Isaac Newton & town       \\
                              \bottomrule   
  \end{tabular}
  }
  \caption{Different verbalized names of the same entity lead to different predictions on BERT-large.}
  \label{tab:sub_var}
\end{table}

\begin{figure}[tp]
  \centering
  \includegraphics[width=0.5\textwidth]{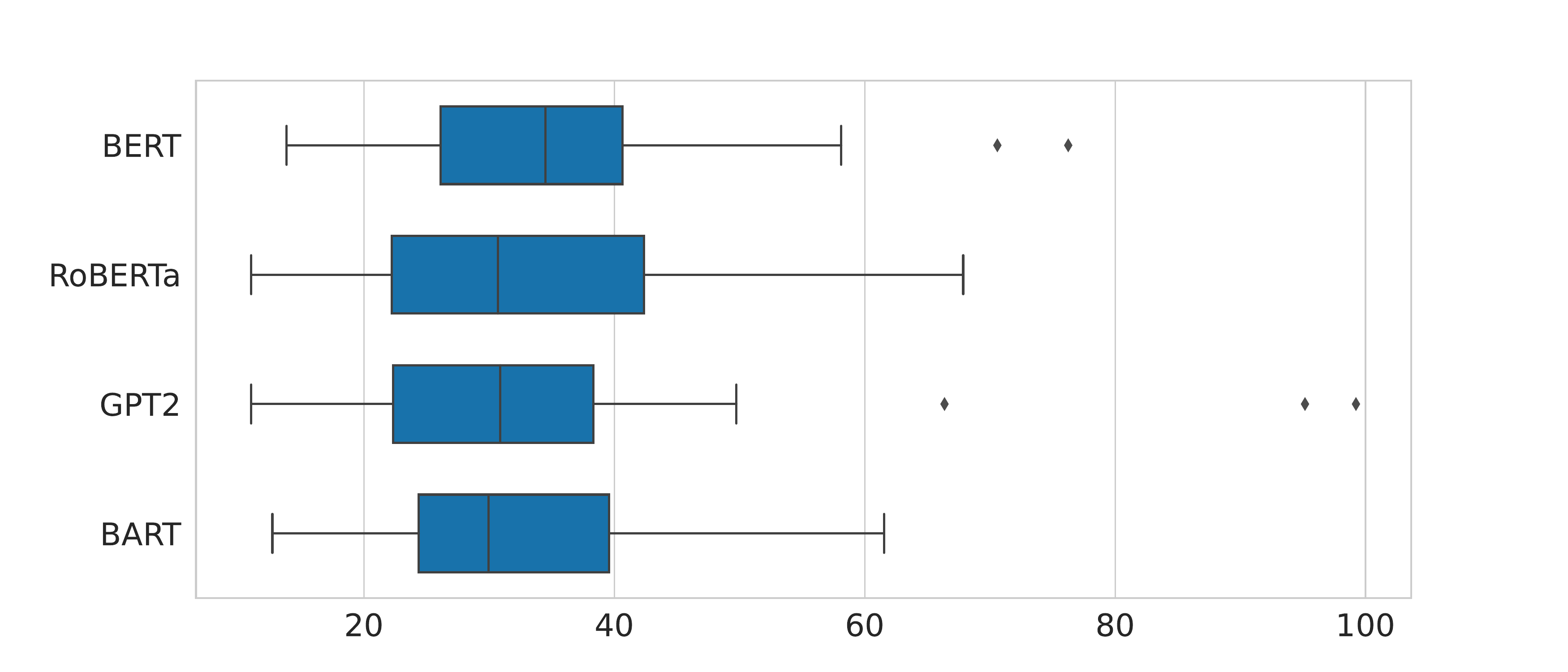}
  \caption{The \emph{verbalization stabilities} of 4 PLMs on all relations, which is measured by the percentage of relation instances whose predictions are unchanged when verbalization varies. We can see that the verbalization stabilities of all 4 PLMs (BERT-large, RoBERTa-large, GPT2-xl, BART-large) are poor.}
  \label{fig:mention_consis}
\end{figure}

To quantify the effect of instance verbalization bias, 
we collect at most 5 verbalizations for each subject entity in LAMA from Wikidata, 
and calculate the \emph{verbalization stability} on each relation, i.e., the percentage of relation instances whose predictions are unchanged when verbalization varies. 
The results in Figure~\ref{fig:mention_consis} show the average verbalization stabilities of all four PLMs are < 40\%, which demonstrate that the instance verbalization bias will bring unstable and unreliable evaluation.

\subsection{Cause of Instance Verbalization Bias}

Figure~\ref{fig:detail_scm}c shows the cause of instance verbalization bias: the backdoor path $M \leftarrow C \leftarrow L \rightarrow X \rightarrow E$,
which stems from the confounder of linguistic distribution $L$  between pretraining corpus $C$ and verbalized probing data $X$. 
Consequently, the observed correlation between $M$ and $E$ couldn't faithfully represent the true causal effect of $M$ on $E$, but is also mixed up the spurious correlation caused by the backdoor path.

The instance verbalization bias can be eliminated by blocking this backdoor path via causal intervention, which requires a distribution formulation of the instance verbalization, i.e., $\mathcal{P}(X)$. We will present a possible intervention formulation in Section~\ref{sec:reduce}.

\section{Sample Disparity Bias}
Besides the biases induced by linguistic correlations, the distributional correlations between pretraining corpus and task-specific probing data can also introduce sample disparity bias. That is, the performance difference between different PLMs may due to the sample disparity of their pretraining corpus, rather than their ability divergence.

In conventional evaluation, the evaluated hypotheses are independent of the train/test data generation, and all the hypotheses are evaluated on training data and test data generated from the same distribution. 
Therefore, the impact of correlations between training data and test data is transparent, controllable, and equal for all the hypotheses.
By contrast, in prompt-based probing, each PLM is bundled with a unique pretraining corpus, the correlation between pretraining corpus distribution and probing data distribution cannot be quantified. In the following we investigate this sample disparity bias in detail.

\subsection{Sample Disparity Brings Biased Performance}
In factual knowledge probing, LAMA~\citep{petroniLanguageModelsKnowledge2019}, a subset sampled from Wikidata, is commonly used to compare different PLMs. Previous work claims that GPT-style models are with weaker factual knowledge extraction abilities than BERT because they perform worse on LAMA~\cite{petroniLanguageModelsKnowledge2019,DBLP:journals/corr/abs-2103-10385}. 
However, because PLMs are pretrained on different pretraining corpus, the performance divergence can stem from the spurious correlation between pretraining corpus and LAMA, rather than their ability difference. 
For example, BERT's superior performance to GPT-2 may stem from the divergence of their pretraining corpus, where BERT's pretraining corpus contains Wikipedia, while GPT-2's pretraining corpus doesn't.

To verify the effect of sample disparity bias, we further pretrain BERT and GPT-2 by constructing pretraining datasets with different correlation degrees to LAMA, and report their new performances on LAMA. Specifically, we use the Wikipedia snippets in LAMA and collect a 99k-sentence dataset, named WIKI-LAMA. Then we create a series of pretraining datasets by mixing the sentences from WIKI-LAMA with WebText\footnote{\url{http://Skylion007.github.io/OpenWebTextCorpus}} (the pretraining corpus of GPT2). That is, we fix all datasets’ size to 99k, and a parameter $\gamma$ is used to control the mixture degree: for each dataset, there are $\gamma\%$ instances sampled from WIKI-LAMA and $1 - \gamma \%$ instances sampled from WebText. Please refer to the appendix for pretraining detail.

 \begin{table}[tp]
 \setlength{\belowcaptionskip}{-0.3cm}
  \resizebox{\columnwidth}{!}{
  \begin{tabular}{ccccc}
  \toprule
    \textbf{$\gamma \%$} & \textbf{BERT-base} & \textbf{BERT-large} & \textbf{GPT2-base} & \textbf{GPT2-medium} \\ \hline
  0\%        &  30.54    &    33.08        &   15.22        &   22.11      \\
  20\%       &  35.77    &    39.56        &   22.02        &   28.21      \\
  40\%       &  38.68    &    39.75        &   24.32        &   30.29      \\
  60\%       &  38.72    &    40.68        &   25.42        &   31.16      \\
  80\%       &  39.79    &    41.48        &   25.65        &   31.88      \\
  100\%      &  40.15    &    42.51        &   26.82        &   33.12        \\ \hline
  None        & 37.13    &    39.08        &   16.88       &   22.60        \\ 
  \bottomrule      
  \end{tabular}
  }
  \caption{The P@1 on LAMA of PLMs whose further pretraining data are with different correlation degrees $\gamma \%$ with LAMA. The BERT-base and GPT2-base both contain 12 layers, while BERT-large and GPT2-medium both contain 24 layers.}
  \label{tab:gamma_change}
  \end{table}

Table~\ref{tab:gamma_change} demonstrates the effect of sample disparity bias. We can see that 
1) Sample disparity significantly influences the PLMs’ performance: the larger correlation degree $\gamma$ will result in better performance for both BERT and GPT-2;
2) Sample disparity contributes to the performance difference. We can see that the performance gap between GPT-2 and BERT significantly narrows down when they are further pretrained using the same data. Besides, further pretraining BERT on WebText ($\gamma$=0) would significantly undermine its performance. 
These results strongly confirm that the sample disparity will significantly bias the probing conclusion.

\subsection{Cause of Sample Disparity Bias}
The cause of sample disparity bias may diverge from PLMs and scenarios due to the different causal relation between pretraining corpus distribution $D_a$ and probing data distribution $D_b$. Nevertheless, sample disparity bias always exist because the backdoor path will be $M \leftarrow C \leftarrow D_a \rightarrow D_b \rightarrow T \rightarrow X\rightarrow E$ when $D_a$ is the ancestor of $D_b$, or $M \leftarrow C \leftarrow D_a \leftarrow D_b \rightarrow T \rightarrow X\rightarrow E$ when $D_a$ is the descendant of $D_b$.
Figure~\ref{fig:detail_scm}d shows a common case when the pretraining corpus distribution $D_a$ is an ancestor of probing data distribution $D_b$. For example, the pretraining data contains Wikipedia and probing data is a sampled subset from Wikipedia (e.g., LAMA, X-FACTR, BioLAMA). As a result, there is a backdoor path between $M$ and $E$, which will mislead the evaluation.

\section{Bias Elimination via Causal Intervention}
\label{sec:reduce}
This section describes how to eliminate the above-mentioned biases by blocking their corresponding backdoor paths. 
According to the Backdoor Criterion in Section~\ref{ssec:causal}, we need to choose a set of variables $Z$ that can block every path containing an arrow into $M$ between $M$ and $E$. 
Since the linguistic distribution $L$, pretraining corpus distribution $D_a$ and probing data distribution $D_b$ are unobservable, we choose $Z = \{P, X\}$ as the variable set for blocking all backdoor paths  between $(M, E)$ in the SCM by conducting backdoor adjustment:
\begin{equation}
  \begin{split}
    &\mathcal{P}(E|do(M=m), R=r) = \\ & \sum_{p \in P} \sum_{x \in X}   \mathcal{P}(p,x) \mathcal{P}(E|m, r, p, x).
  \end{split}
  \label{eq:backdoor}
\end{equation}

Equation~\ref{eq:backdoor} provides an intuitive solution. To eliminate the biases stemming from the spurious  correlations between pretraining corpus, probing data and prompts, we need to consider the natural distribution of prompts and verbalized probing data regardless of other factors. 
Consequently, the overall causal effects between PLM and evaluation result are the weighted averaged effects on all valid prompts and probing data.

Unfortunately, the exact distribution of $\mathcal{P}(x,p)$ is intractable , which needs to iterate over all valid prompts and all verbalized probing data. 
To address this problem, we propose a sampling-based approximation.
Specifically, given a specific assumption about $\mathcal{P}(x,p)$ (we assume uniform distribution in this paper without the loss of generality), we sample $K_p$ prompts for each relation and $K_x$ kinds of verbalization for each instance according to $\mathcal{P}(x,p)$, and then these samples are used to estimate the true causal effects between $M$ and $E$ according to Equation~\ref{eq:backdoor}.

To verify whether causal intervention can improve the evaluation consistency and robustness, we conduct backdoor adjustment experiments on 8 different PLMs. 
We randomly sample 1000 subsets with 20 relations from LAMA, and observe whether the evaluation conclusions were consistent and stable across the 1000 evaluation runtimes.
Specifically, we use \textbf{\textit{rank consistency}} as the evaluation metric, which measures the percentage of the most popular rank of each model in 1000 runtimes. For example, if BERT ranks at $3^{rd}$ place in 800 of the 1000 runtimes, then the rank consistency of BERT will be $80\%$.

\begin{table}[tp]
\setlength{\belowcaptionskip}{-0.3cm}
\centering
\resizebox{\columnwidth}{!}{
\begin{tabular}{lccc}
\toprule
\textbf{Model} &
\textbf{Original} &
\textbf{Random} &
\textbf{+Intervention} \\ \hline
BERT-base     & 56.4 & 45.4   & \textbf{86.5}   \\
BERT-large    & 100.0 & 78.1   & \textbf{100.0}  \\
RoBERTa-base  & 75.7 & 44.0   & \textbf{77.8}   \\
RoBERTa-large & 56.1 & 42.2   & \textbf{86.5}   \\
GPT2-medium   & 63.5 & 40.7   & \textbf{98.2}  \\
GPT2-xl       & 74.2 & 35.7   & \textbf{77.8}    \\
BART-base     & 63.4 & 61.6   & \textbf{98.2}  \\
BART-large    & 97.7 & 61.3  & \textbf{100.0}  \\ \hline
Overall Rank     & 25.5 & 5.5    & \textbf{68.5}  \\ \bottomrule
\end{tabular}
}
\caption{The \emph{rank consistencies} over 1000 task samples (each task contains 20 relations from LAMA). For a PLM, the rank consistency is the percentage of its most popular rank in 1000 runtimes. For ``Overall Rank'', the rank consistency is the percentage of the most popular rank of all PLMs in 1000 runtimes, i.e., the rank of all PLMs remains the same. ``Original'' means that we use the LAMA's original prompts and verbalized names, ``Random'' means that we randomly sample prompts and verbalized names every time, ``+Intervention'' means that we apply causal intervention. We can see that the rank consistency is signiﬁcantly raised after causal intervention.}
\label{tab:rank_cons}
\end{table}

Table~\ref{tab:rank_cons} shows the results. We can see that causal intervention can significantly improve the evaluation consistency:
1) The consistency of current prompt-based probing evaluations is very poor on all 8 PLMs: when we randomly select prompts and verbalizations during each sampling, the overall rank consistency is only 5.5\%;
2) Causal intervention can significantly improve overall rank consistency: from 5.5\% to 68.5\%;
3) Casual intervention can consistently improve the rank consistency of different PLMs: the rank of most PLMs is very stable after backdoor adjustment.

The above results verify that causal intervention is an effective technique to boost the stability of evaluation, and reach more consistent conclusions.

\section{Related Work}

\paragraph{Prompt-based Probing} 
Prompt-based probing is popular in recent years~\citep{rogersPrimerBERTologyWhat2020,liu2021pre} for probing factual knowledge~\citep{petroniLanguageModelsKnowledge2019,jiangXFACTRMultilingualFactual2020,sung2021can}, commonsense knowledge~\citep{feldmanCommonsenseKnowledgeMining2019}, semantic knowledge~\citep{ettingerWhatBERTNot2020,DBLP:journals/corr/abs-2107-02137,brownLanguageModelsAre2020,schickFewShotTextGeneration2020} and syntactic knowledge~\citep{ettingerWhatBERTNot2020} in PLMs.  
And a series of prompt-tuning studies consider optimizing prompts on training datasets with better performance but may undermine interpretability~\citep{jiangHowCanWe2020,shinAutoPromptElicitingKnowledge2020,haviv-etal-2021-bertese,gaoMakingPretrainedLanguage2020,qin2021learning,liPrefixTuningOptimizingContinuous2021,zhong2021factual}. Because such prompt-tuning operations will introduce additional parameters and more unknown correlations, this paper does not take prompt-tuning into our SCM, delegate this to future work.

\paragraph{Biases in NLP Evaluations} 
Evaluation is the cornerstone for NLP progress. 
In recent years, many studies aim to investigate the underlying biases and risks in evaluations. 
Related studies include investigating inherent bias in current metrics~\citep{DBLP:conf/mtsummit/Coughlin03,callison2006re,DBLP:conf/emnlp/LiMSJRJ17,DBLP:conf/aaai/SaiGKS19,DBLP:journals/corr/abs-2008-12009}, 
exploring dataset artifacts in data collection and annotation procedure~\citep{lai-hockenmaier-2014-illinois,marelli-etal-2014-semeval,DBLP:conf/acl/InkpenZLCW18,levy-dagan-2016-annotating,schwartz-etal-2017-effect,cirik2018visual,mccoy-etal-2019-right,DBLP:conf/emnlp/0001BPRCE21, branco-etal-2021-shortcutted},
and identifying the spurious correlations between data and label which might result in catastrophic out-of-distribution robustness of models~\citep{DBLP:conf/emnlp/PoliakHRHPWD18,DBLP:conf/naacl/RudingerNLD18,DBLP:conf/acl/SmithCSRA18}.

Most previous studies demonstrate the evaluation biases empirically, and interpret the underlying reasons intuitively. However, intuitive explanations are also difficult to critical and extend.
In contrast, this paper investigates the biases in prompt-based probing evaluations from a causal view. Based on the causal analysis framework, we can identify, understand, and eliminate biases theoretically, which can be extended and adapted to other evaluation settings in a principled manner\footnote{Greatly inspired by the reviewer's valuable comments.}.
We believe both the causal analysis tools and the valuable insights can benefit future researches.

\section{Conclusions and Discussions} 
This paper investigates the critical biases and quantifies their risks in the widely used prompt-based probing evaluation, including prompt preference bias, instance verbalization bias, and sample disparity bias.
A causal analysis framework is proposed to provide a unified framework for bias identification, interpretation and elimination with a theoretical guarantee.
Our studies can promote the understanding of prompt-based probing, remind the risks of current unreliable evaluations, guide the design of unbiased datasets, better probing frameworks, and more reliable evaluations, and push the bias analysis from empirical to theoretical.

Another benefit of this paper is to remind the evaluation criteria shifts from conventional machine learning algorithms to pretrained language models. As we demonstrate in Figure~\ref{fig:head}, in conventional evaluation, the evaluated hypotheses (e.g., algorithms, architectures) are raised independently of the train/test dataset generation, where the impact of correlations between training data and test data is transparent, controllable, and equal for all the hypotheses. However, in evaluations of pretrained language models, the pretraining corpus is bundled with the model architecture. In this case, it is significant to distinguish what you need to assess (architecture, corpus, or both), as well as the potential risks raised by the correlations between pretraining corpus and test data, which most current benchmarks have ignored. Consequently, this paper echoes that it is necessary to rethink the criteria for identifying better pretrained language models, especially under the prompt-based paradigm.

In the future, we would like to extend our causal analysis framework to fit prompt-tuning based probing criteria and all PLM-based evaluations.

\section*{Acknowledgments}
We sincerely thank all anonymous reviewers for their insightful comments and valuable suggestions. This research work is supported by the National Natural Science Foundation of China under Grants no. 62122077, the Strategic Priority Research Program of Chinese Academy of Sciences under Grant No. XDA27020200, and the National Natural Science Foundation of China under Grants no. 62106251 and 62076233.

\section*{Ethics Consideration}
This paper has no particular ethic consideration.

\bibliography{anthology,custom,updated}
\bibliographystyle{acl_natbib}

\newpage

\appendix

\section{Datasets Construction Details}
\paragraph{Instance Filtering} We follow the data construction criteria as LAMA, we remove the instances whose object is multi-token or not in the intersection vocabulary of these 4 PLMs. 

\paragraph{Relation Selection} We remove all the N-M relations in LAMA such as ``share border with'' or ``twin city''. Because in these relations, there are multiple object entities corresponding to the same subject entity. In that case, the metric Precision@1 is not suitable for evaluating PLMs in such relations. In addition, due to the completeness limitation of knowledge bases, it's impossible to find all the correct answers for each subject. Therefore, we do not include these relations in our experiments.

\paragraph{Prompt Generation} Because of the difference between the pretraining tasks of these 4 PLMs (autoencoder, autoregressive and denoising autoencoder), we design prompts where the placeholder for the target object is at the end, e.g., \textit{The birthplace of $x$ is $y$} instead of \textit{$y$ is the birthplace of $x$}.
We follow the instruction from Wikidata, combine the prompts from~\citet{elazar2021measuring} and~\citet{jiangHowCanWe2020}, and manually filter out the prompts with inappropriate semantics. 
All the prompts are created before any experiments and fixed afterward.

\section{Further Pretraining Details}

We further pretrain BERT with masked language modeling (mask probability=$15\%$) and GPT2 with autoregressive language modeling task respectively. Training was performed on $8$ 40G-A100 GPUs for $3$ epochs, with maximum sequence length $512$. The batch sizes for BERT-base, BERT-large, GPT2-base, GPT2-medium are $256, 96, 128, 64$ respectively. All the models is optimized with Adam using the following parameters: $\beta_1=0.9, \beta_2=0.999, \epsilon=1e-8$ and the learning  rate is $5e-5$ with warmup ratio=$0.06$. 

\end{document}